# Belief Induced by the Partial Knowledge of the Probabilities

Philippe Smets
IRIDIA
Université Libre de Bruxelles
Brussels-Belgium

### Abstract:

We construct the belief function that quantifies the agent' beliefs about which event of $\Omega$ will occurred when he knows that the event is selected by a chance set-up and that the probability function associated to the chance set up is only partially known.

## 1. INTRODUCTION.

1) The use of belief functions to quantify degrees of belief is muddled by problems that result from the confusion between belief and lower probabilities (or between plausibility and upper probabilities). Beliefs can be induced by many types of information. In this paper, we consider only one very special case: beliefs induced on a frame of discernment $\Omega$ when the elements of $\Omega$ will be selected by a random process. It seems reasonable to defend the idea that the belief of an event should be numerically equal to the probability of that event. This principle is called the Hacking Frequency Principle (Hacking 1965).

But there are cases where the probability function that governs the random process is not exactly known. This lack of knowledge can be encountered when probabilities are partially defined or when data are missing. As an example, suppose an urn where there are 100 balls. Its composition is not exactly known. All that is known is that there are between 30 and 40 black balls, between 10 and 50 white balls, and the other are red. What is your belief that the next randomly selected ball will be black? Suppose you have selected 50 balls at random with replacement and you have observed 15 black balls, 20 white, 10 reds and 5 'not black'. What is your belief now that there are between 35 and 37 black balls? What is your belief now that the next randomly selected ball will be black? These are the problems we solve in this paper.

In this paper, we accept that beliefs are quantified by belief functions, as described in the transferable belief model (Smets 1990b, Smets and Kennes 1994). The transferable belief model is a model for quantified beliefs developed independently of any underlying probabilistic model. It is neither Dempster's model nor its today versions (Shafer, 1990, Kohlas, 1994). It is not a model based on inner measures (Halpern and Fagin, 1990).

What we study here is just a special case of belief function. We study the belief induced by the knowledge of the existence of an objective chance set up that generates random events according to a probability function, probability function that happens to be only partially known to us.

2) Suppose a frame of discernment $\Omega$, i.e., a set of mutually exclusive and exhaustive events such as one and only one of them is true (we accept the close world assumption (Smets 1988)). Suppose the true element will be selected by a chance process. Let $P:2^\Omega \to [0,1]$ be the probability function over $\Omega$ where $P(A)$ for $A \subseteq \Omega$ quantifies the probability (chance) that the selected element is in A. We accept that this probability measure is "objective". The problem is to assess Your degree of belief. You denotes the agent who hold the beliefs. Your beliefs are quantified by a belief function $bel:2^\Omega \to [0,1]$, about the fact that the selected element is in A, given You only have some partial knowledge about the value of P.

Should You know P, then by Hacking Frequency Principle (1965) Your degree of belief bel(A) for each $A \subseteq \Omega$ should be equal to P(A):
If You know that $P(A) = p_A \; \forall A \subseteq \Omega$
$$\text{then bel}(A) = p_A \; \forall A \subseteq \Omega$$
In that case bel is a probability function over $\Omega$. But remember that bel and P do not have the same meaning; they only share the same values. P quantifies the probability (chance) of the events in $\Omega$, bel quantifies the belief over $\Omega$ induced in You by the knowledge of the value of the probabilities. P exists independently of me; bel cannot exist if You do not exist.

Let $\mathbb{P}_\Omega$ be the set of probability functions over $\Omega$. Suppose that You know only that the probability function P that governs the random process over $\Omega$ is an element of a subset $\mathcal{P}$ of $\mathbb{P}_\Omega$. The problem is to determine Your belief about $\Omega$ given You know only that P is an element of $\mathcal{P}$ (but You do not know which one).

In many cases, $\mathcal{P}$ is uniquely defined by its upper and lower probabilities functions $P^*$ and $P_*$ where:
$$P_*(A) = \min \{ P(A) : P \in \mathcal{P} \}$$

$$P^*(A) = \max \{ P(A) : P \in \mathcal{P} \} = 1 - P_*(\overline{A})$$

or $\mathcal{P} = \{P : P \in \mathbb{P}_\Omega, P_*(A) \leq P(A) \leq P^*(A), \forall A \subseteq \Omega\}$.

Just as P and bel characterize different concepts, $P_*$ and bel characterize also different concepts, even when $P_*$ is mathematically a belief function. The function bel concerns Your belief over $\Omega$. The function $P_*$ gives the lowest possible values for the probability of the events in $\Omega$ compatible with what You know.

This knowledge that $P \in \mathcal{P} \subseteq \mathbb{P}_\Omega$ is translated into a belief $bel_{\mathbb{P}_\Omega}$ over $\mathbb{P}_\Omega$.[1] That belief only supports $\mathcal{P}$, i.e., its basic belief masses are:

$$m_{\mathbb{P}_\Omega}(\mathcal{Q}) = 1 \quad \text{if } \mathcal{Q} = \mathcal{P}$$
$$= 0 \quad \text{otherwise}$$

Given Your belief over $\mathbb{P}_\Omega$, can You build Your belief over $\Omega$. In this paper, we will show how to build such a belief function.

Classical material about belief functions and the transferable belief model can be found in Shafer (1976), Smets (1988) and Smets and Kennes (1994).

## 2. IMPACT OF HACKING FREQUENCY PRINCIPLE.

The general frame consists of:
- $\Omega$: the finite set of possible elementary events $\omega_i$, $i=1, 2...n$, (the outcomes of the stochastic experiment);
- $\mathbb{P}_\Omega$: the set of probability functions P over $\Omega$;
- $\mathbb{B}_{\mathbb{P}_\Omega}$: the set of belief functions over $\mathbb{P}_\Omega$.

Let $N = \{1, 2...n\}$. Let $W = \mathbb{P}_\Omega \times \Omega$. All subsets A of W can be represented as the finite union of the intersection of A with each of the elementary events $\omega_i$:

$$\forall A \subseteq W, \quad A = \bigcup_{i \in N} (A_i, \omega_i) \quad (2.1)$$

where $A_i = proj(A \cap cyl(\omega_i)) \subseteq \mathbb{P}_\Omega$, cyl(X) is the cylindrical extension of X on W where X denotes a subset of $\Omega$ (or $\mathbb{P}_\Omega$), and proj(B) is the projection of $B \subseteq W$ on $\Omega$ (or $\mathbb{P}_\Omega$) (context makes it clear which domain and which range are involved).

The major problem solved in this paper is the construction of the belief function $bel_W$ on W that would result if You were in a state of total ignorance about the value of P. If You have some prior belief $bel_{\mathbb{P}_\Omega}$ about the value of P, the belief $bel_W$ over W would be combined with the vacuous extension of $bel_{\mathbb{P}_\Omega}$ on W by the application of Dempster's rule of combination. We will treat essentially the case where You only know that $P \in \mathcal{P}$ where $\mathcal{P}$ is a subset of $\mathbb{P}_\Omega$, i.e., when $\mathcal{P}$ is the only focal element and Your belief over $\mathbb{P}_\Omega$ can be represented by the basic belief assignment with $m_{\mathbb{P}_\Omega}(\mathcal{P}) = 1$. Generalization for a finite (or countable) numbers of focal elements is immediate. Further generalization is more delicate.

Let $\mathbb{B}_W$ be the set of belief functions over W. What is their nature? We are going to construct the equivalent of the basic belief masses (bbm) on $\mathbb{B}_W$. We say equivalent as W is not a finite space and the concept of basic belief masses has to be extended in order to cope with the structure of W. The bbm will become some sort of 'densities'. For simplicity sake, they are also denoted by $m_W : 2^W \to [0,1]$. The value $bel_W(A)$ is defined as the 'integral' of the $m_W$ values given to the non empty subsets of A. It happens that in the case considered in this paper, $m_W$ is a real density for which classical integrals are well defined. We call the $m_W$ function a basic belief density (bbd) to enhance its particular nature. Those subsets A of W such that $m_W(A)>0$ are called the focal elements of $m_W$.

The first constraint about $m_W$ results from Hacking Frequency Principle. Suppose You know the values $P(\omega_i)$ of the objective probability function P on $\Omega$ for every $\omega_i \in \Omega$ (what is translated by $\mathcal{P} = \{P\}$). Let $bel_\Omega^{\{P\}}$ denotes Your belief over $\Omega$ when You know that $\mathcal{P} = \{P\}$.[2] By Hacking Frequency Principle, the value $bel_\Omega^{\{P\}}(X)$ for any subset X of $\Omega$ is numerically equal to the probability P(X) given to X.

By construction, $bel_\Omega^{\{P\}}$ results from the marginalizing of $bel_W^{\{P\}}$ over $\Omega$:

$$bel_\Omega^{\{P\}}(X) = bel_W^{\{P\}}(cyl(X)) \quad \forall X \subseteq \Omega$$

Hacking Frequency Principle implies the next requirement.

**Requirement 1:**
If You know that $\mathcal{P} = \{P\}$
then $bel_W^{\{P\}}(cyl(X)) = P(X) \quad \forall X \subseteq \Omega \quad (2.2)$

Let $\pi(P) = cyl(\{P\}) \subseteq W$, then $\pi(P) = \bigcup_{i \in N} (\{P(\omega_i)\}, \omega_i)$ by 2.1. Thus $bel_W^{\{P\}}(cyl(X)) = bel_W(cyl(X)|\pi(P))$. The second term is just the result of the conditioning of $bel_W$ on $cyl(\{P\})$, what is achieved by the application of Dempster's rule of conditioning. Hence the bbd $m_W(A)$, $A \subseteq W$ is transferred to $A \cap \pi(P)$. Let $A = \bigcup_{i \in N} (A_i, \omega_i) \subseteq W$, then m(A) is transferred to $A \cap \pi(P) = \bigcup_{i \in N} (A_i \cap \{P(\omega_i)\}, \omega_i)$.

---

[1] subscripts of m and bel denote their domain.

[2] Superscripts of bem and m denote Your knowledge about P, i.e., the focal element of $bel_{\mathbb{P}_\Omega}$.

The result of this conditioning on $\pi(P)$ is a probability function. Hence $\text{bel}_W^{\{P\}}$ must be a Bayesian belief function, i.e., only singletons can be focal elements and $\text{bel}_W^{\{P\}}(W) = 1$. The singletons of W have the form $(\{P(\omega_i)\}, \omega_i)$ for $i \in \{1,...n\}$, $P \in \mathbb{P}_\Omega$. Hence $m_W$ must satisfy:

$$m_W(\bigcup_{i \in N}(A_i, \omega_i)) = 0 \quad \text{if } A_i \cap A_j \neq \emptyset \text{ for some } i \neq j \in N$$
$$= 0 \quad \text{if } \bigcup_{i \in N} A_i \neq \mathbb{P}_\Omega$$
$$\geq 0 \quad \text{otherwise.} \quad (2.3)$$

The impact of Hacking Frequency Principle, translated by 2.3, is very strong. It implies that the focal elements of $m_W$ can be represented as $\bigcup_{i \in N}(A_i, \omega_i)$ where the $A_i$, i=1,...n, are non empty elements of a partition of $\mathbb{P}_\Omega$.

## 3. THE CASE WHERE $|\Omega| = 2$.

We study now the case where $|\Omega| = 2$. Let $\Omega = \{S, F\}$ where S and F denote Success and Failure, respectively.

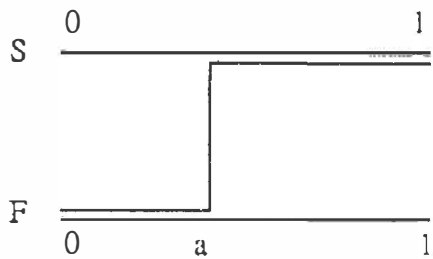

Figure 1: Structure of the domain of $m_W$ and one example of bbd centered on a when $|\Omega| = 2$.

Let $A \subseteq W$ be a focal element of $m_W$, then $A = (\alpha, S) \cup (\overline{\alpha}, F)$ where $\alpha \subseteq [0, 1]$ and $\overline{\alpha}$ is the complement of $\alpha$ relative to [0, 1]. For simplicity sake, the bbd $m_W((\alpha, S) \cup (\overline{\alpha}, F))$ is written as $m_W(\alpha)$.

W can be graphically represented by two [0, 1] intervals where the upper [0, 1] interval is the intersection of W with S, and the lower one is the intersection of W with F (see figure 1). Every focal element of $m_W$ is made of a set of mutually exclusive and exhaustive intervals that are either in the S domain or the F domain. By convention, intervals are defined as closed to the left and open to the right, except when 1 is the right limit, in which case the interval is also closed to the right.

We introduce an extra assumption.

**Requirement 2:**
If $|\Omega|=2$, $\text{bel}_{\mathbb{P}_\Omega}([a,b) \cup [c,d) | S) = \text{bel}_{\mathbb{P}_\Omega}([a,d) | S)$
for every $0 \leq a \leq b \leq c \leq d \leq 1$.

It is equivalent to assuming that $m_W(\alpha)$ is null except if $\alpha = [a, 1]$. The origin of the assumption is to be found in the meaning of the bbd. The bbd $m_W(\alpha)$ for $\alpha \subseteq [0, 1]$ is that part of belief (a density here) that supports the fact that $P(S) \in \alpha$ (and $P(F) \in \overline{\alpha}$). Suppose we condition $m_W$ on S. Each bbd $m_W(\alpha)$ is transferred to $(\alpha, S) \subseteq W$. Requirement 2 means that if after conditioning on S a bbd supports $P(S) = x \in [0,1]$, it also supports every value in [0, 1] larger than x. Observing a success could support $P(S) = .3$, but that support should then also be given to $P(S) = .4$ etc....

This assumption means that each focal element is a step function that starts from $(\{0\}, F)$, jumps from the F domain to the S domain at some a in [0,1], and ends at $(\{1\}, S)$ (see figure 1).

Finally, if we apply again the Hacking Frequency Principle, we obtain after conditioning on $\mathcal{P} = \{P\}$ with $P(S) = p$, $P(F) = 1-p$:

$$\text{bel}_\Omega^{\{P\}}(S) = p = \int_0^p m_W([x,1]) \, dx.$$

The second equality results from the fact that only those bbd that jump before p will touch $(\{p\}, S)$ and $\text{bel}_\Omega^{\{P\}}(S)$ is equal to the integral of those bbd that touch $(\{p\}, S)$. Derivating both terms on p implies that:
$$m_W([p,1]) = 1 \quad \forall p \in [0,1].$$

In conclusion we have derived the bbd on $W$ when $|\Omega| = 2$.

Some properties can be easily derived.

1) Suppose the agent knows that $\mathcal{P} = \{P: a \leq P(S) \leq b, 0 \leq a < b \leq 1\}$. We condition $m_W$ on the cylindrical extension of [a, b]. The bbd $m_W(A)$ for $A \subseteq W$ is transferred to $A \cap \text{cyl}([a, b])$. $\text{bel}_W^\mathcal{P}(S)$ is the integral of all the bbd that touch only S after conditioning on $\text{cyl}([a, b])$, i.e., those bbd that jump to S before a:

$$\text{bel}_W^\mathcal{P}(S) = \int_0^a m_W([x,1]) \, dx = a$$

Similarly $\text{pl}_W^\mathcal{P}(S)$ is the integral of all the bbd that touch S, i.e., that jump to S before b:

$$\text{pl}_W^\mathcal{P}(S) = \int_0^b m_W([x,1]) \, dx = b$$

This result should not be extrapolated blindly to higher dimensions (see section 4).

2) The case $|\Omega| = 2$ can be nicely represented by figure 2 (Smets, 1978). Each point in the triangle corresponds to one interval of [0,1]. In general, if positive bbd are given only to intervals, we assign the bbd given to [a,b] to the point (a,b) of the triangle. Then:

$$\text{bel}([a,b]) = \int_a^b \int_x^b m([x,y])\, dy\, dx$$

$$\text{pl}([a,b]) = \int_0^b \int_{a \vee x}^1 m([x,y])\, dy\, dx$$

$$q([a,b]) = \int_0^a \int_b^1 m([x,y])\, dy\, dx$$

$$m([a,b]) = -\frac{\partial^2 \text{bel}([a,b])}{\partial a\, \partial b} = -\frac{\partial^2 q([a,b])}{\partial a\, \partial b}$$

The result of the application of Dempster's rule of combination is given by multiplying the commonality functions.

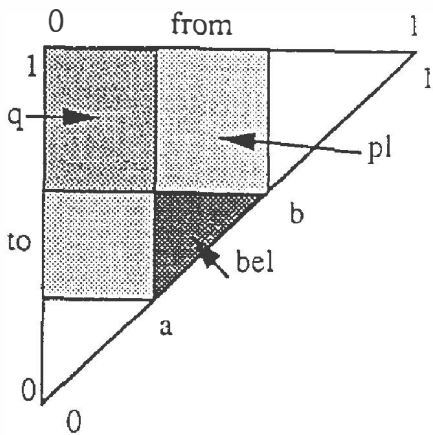

**Figure 2:** Parametric representation on beliefs on [0,1] when the focal elements are intervals. The shaded areas are those on which integration is performed in order to compute bel([a,b]) (a triangle), q([a,b]) (a rectangle) and pl([a,b]) (a rectangle with right lower corner truncated).

In the present case ($|\Omega|=2$) the non-null bbd of $m_W$ obtained after conditioning on S are given to the intervals [a,1], hence they cluster on the upper horizontal line. Those obtained after conditioning on F are given to the intervals [0,b], hence they cluster on the left vertical line.

Suppose You perform n independent experiments and observe r successes, s failures where $r + s \leq n$ (the difference n - (r + s) is the number of experiments for which the outcome is not available). The commonality function induced on $\mathbb{P}_\Omega = [0, 1]$

- by a success is: $\quad q_{\mathbb{P}_\Omega}([a,b] \mid S) = a$
- by a failure is: $q_{\mathbb{P}_\Omega}([a,b] \mid F) = 1-b$
- by a 'S∪F' is: $q_{\mathbb{P}_\Omega}([a,b] \mid S \cup F) = 1$

The belief function induced by 'S∪F' is the vacuous belief function that reflect the state of total ignorance in which You are after just learning the tautology 'S∪F'. Hence we can just as well drop all 'vacuous' results and assume n = r+s.

The commonality function induced by r successes and s failures in n independent (Bernoullian) trials is obtained by multiplying the corresponding commonality functions. Hence:

$$q_{\mathbb{P}_\Omega}([a,b] \mid r, s) = a^r (1-b)^s$$

In that case, by derivating $q_{\mathbb{P}_\Omega}([a,b] \mid r, s)$ and appropriate normalization, we get:

$$m_{\mathbb{P}_\Omega}([a,b] \mid r, s) = \frac{\Gamma(r+s+1)}{\Gamma(r)\, \Gamma(s)} a^{r-1} (1-b)^{s-1}$$

where $\Gamma$ is the gamma function.

When $n \to \infty$, $r \to np$, $s \to n(1-p)$ (hence $p = \lim \frac{r}{r+s}$), the limit of $m([a,b] \mid r, s)$ tends to 0 except for a dirac function at p. In that case $\text{bel}(A \mid r,s) = 1$ if $p \in A$ and 0 otherwise. After accumulating an infinite number of information, You will be in a state of 'total certainty', of 'knowledge' about the value of P(S).

3) Suppose You want to compute the belief that the next outcome is a success (or a failure) given You have already observed r successes and s failures in n independent trials. We use $m([a,b] \mid r, s)$ as the a priori belief over [0, 1]. Dempster's rule of combination $m_{12} = m_1 \oplus m_2$ can be represented as (Dubois and Prade, 1986, Smets, 1993a):

$$m_{12}(A) = \sum_{B \subseteq \Omega} m_1(A \mid B)\, m_2(B)$$

$$\text{bel}_{12}(A) = \sum_{B \subseteq \Omega} \text{bel}_1(A \mid B)\, m_2(B)$$

where $m_1(A \mid B)$ and $\text{bel}_1(A \mid B)$ are unnormalized conditional basic belief masses and belief functions. Generalizing this relation in the present context and denoting $\text{bel}_\Omega\{P: P(S) \in [a,b]\}(S)$ by $\text{bel}_\Omega(S \mid P(S) \in [a,b])$ (which value equals a), one obtains:

$$\text{bel}_\Omega(S \mid r,s) = \int_0^1 \int_a^1 \text{bel}_\Omega(S \mid P(S) \in [a,b])\, m_{\mathbb{P}_\Omega}([a,b] \mid r,s)\, db\, da$$

$$= \int_0^1 \int_a^1 a\, \frac{\Gamma(r+s+1)}{\Gamma(r)\, \Gamma(s)} a^{r-1} (1-b)^{s-1}\, db\, da$$

So: $\quad \text{bel}_\Omega(S \mid r, s) = \dfrac{r}{r+s+1}$

$\text{bel}_\Omega(F \mid r, s) = \dfrac{s}{r+s+1}$

and $\quad m_\Omega(S \cup F \mid r,s) = \dfrac{1}{r+s+1}$.

This result shows that the observed proportion is an excellent approximation of $bel_\Omega$ if $r+s$ is not too small.

## 4. CASE WITH $|\Omega| = 3$.

Suppose $|\Omega| = 3$ where $\Omega = \{A, B, C\}$. $\mathbb{P}_\Omega$ can be represented by an equilateral triangle where each point corresponds to an element of $\mathbb{P}_\Omega$. The three heights are equal to the three probabilities $P(A)$, $P(B)$ and $P(C)$. The height of such a triangle is 1 and the length of its side is equal to $\sqrt{4/3}$.

By requirement 1, we know that the focal elements of $m_W$ can be represented by:

$$(\mathcal{P}_A, A) \cup (\mathcal{P}_B, B) \cup (\mathcal{P}_C, C)$$

where $(\mathcal{P}_A, \mathcal{P}_B, \mathcal{P}_C)$ are the elements of a partition of $\mathbb{P}_\Omega$.

In order to specify the form of the subsets $\mathcal{P}_X$, $X \in \{A, B, C\}$, we consider the conditioning of $m_W$ on the set $\mathcal{P}_L \subseteq \mathbb{P}_\Omega$ where

$$\mathcal{P}_L = \{P: P = (p_A, p_B, p_C): p_B = b_0 + \frac{1 - a_1 - b_0}{a_1} p_A,$$
$$p_C = 1 - p_A - p_B\}. \quad (4.1)$$

where $b_0, a_1 \in [0, 1]$, $b_0 < 1-a_1$.

This set $\mathcal{P}_L$ corresponds to the subset of $\mathbb{P}_\Omega$ where $P(A) \in [0, a_1]$ and $P(B)$ and $P(C)$ are linearly related to $P(A)$. Requirement 3 states that, after conditioning $m_W$ on $\mathcal{P}_L$, the bbd so obtained on the space $\mathcal{P}_L$ is identical to those obtained when $|\Omega|=2$ (indeed every element of the new subdomain is characterized by $P(A)$ as when $|\Omega|=2$). Therefore after further conditioning on A, the focal elements on $\mathcal{P}_L$ should be of the form of intervals $[a, a_1]$ (see figure 3). This requirement is sufficient in order to derive the structure of the focal elements of $m_W$.

**Requirement 3:**
If $|\Omega| = 3$, for every $\mathcal{P}_A$, there exists an $\alpha \in [0, a_1]$ such that the projection of $\mathcal{P}_A$ on $\mathcal{P}_L$ is the interval $[\alpha, a_1]$.

Requirement 3 can identically be defined as:

**Requirement 3':**
If $|\Omega| = 3$, $m_W(X \mid \mathcal{P}_L) \geq 0$ if it exists an $\alpha \in [0, a_1]$
  and $X = \{P: P \in \mathcal{P}_L, p_A \geq \alpha\}$.
  $= 0$ otherwise.

Each requirement implies that the limits between $\mathcal{P}_A$ and $\mathcal{P}_B$ must be a straight line passing through the corner where $P(C) = 1$ and that crosses the opposite side of the triangle (and similarly for the other limits). Every focal elements of $m_W$ can be labeled by an element $q = (q_A, q_B, q_C) \in \mathbb{P}_\Omega$. The focal element labeled by q is the set
$$\mathcal{P}(q) = (\mathcal{P}_A(q), A) \cup (\mathcal{P}_B(q), B) \cup (\mathcal{P}_C(q), C) \quad (4.2)$$
where:
$$\mathcal{P}_A(q) = \{P: P = (p_A, p_B, p_C) \in \mathbb{P}_\Omega,$$
$$p_A \geq \max(p_B \frac{q_A}{q_B}, p_C \frac{q_A}{q_C})\}$$

and similarly for $\mathcal{P}_B(q)$ and $\mathcal{P}_C(q)$ where the A,B,C-indexes are symmetrically exchanged. The graphical representation of $\mathcal{P}_A(q)$ is the upper corner of the $\mathbb{P}_\Omega$ triangle that includes all points in $\mathbb{P}_\Omega$ between the upper corner and the two straight lines drawn from the two other corners through q. $bel_W(X \mid A)$ for $X \subseteq \mathbb{P}_\Omega$ will be the 'integral' of all the bbd given to the focal elements $\mathcal{P}(q)$ such that $X \subseteq \mathcal{P}_A(q)$.

Figure 4 shows the structure of the partition so generated.

The value of $bel_\Omega(A \mid \mathcal{P} = \{(a, b, c)\})$ is the 'integral' of $m_W$ taken over all q in the triangle which corners are (0, 0, 1), (0, 1, 0) and (a, b, c). Reapplying the Hacking Frequency Principle we have:
$$bel_\Omega(A \mid \mathcal{P} = \{(a, b, c)\}) = a \quad (4.3)$$

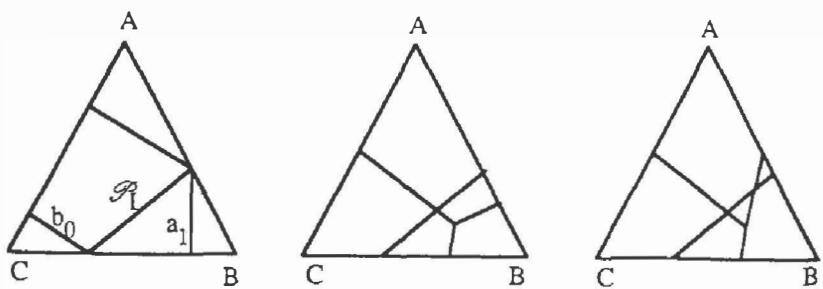

**Figure 3:** Explanation of Requirement 3. Left figure: the $\mathbb{P}_\Omega$ space with $|\Omega| = 3$, and the $\mathcal{P}_L$ domain. Middle figure: a bbd that satisfies Requirement 3. Right figure: a bbd that does not satisfy Requirement 3

It can then be proved that the only function $m_W$ symmetric in the three arguments of q that satisfies (4.3) for every $(a, b, c) \in \mathbb{P}_\Omega$ is the function $m_W(\mathcal{P}(q)) = \sqrt{3}$ for every $q \in \mathbb{P}_\Omega$.

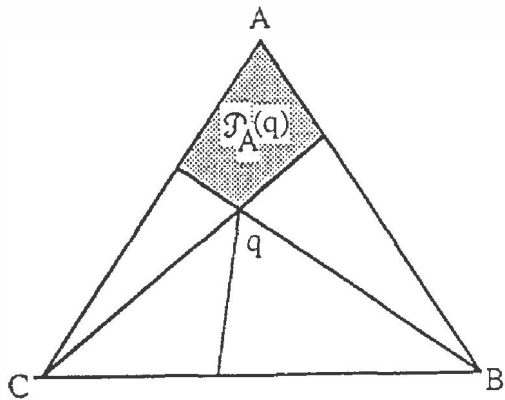

Figure 4: Structure of the domain of $m_W$ and one example of bbd labeled q when $|\Omega| = 3$. The shaded area is $\mathscr{P}_A(q)$.

Some properties derived from this solution are detailed.

1) If $\mathscr{P} = \{(.5, 0, .5), (.5, .5, 0)\}$, then $m_\Omega(A) = 1/3$, $m_\Omega(B) = m_\Omega(C) = 0$, $m_\Omega(A\cup B) = m_\Omega(A\cup C) = 1/6$, $m_\Omega(B\cup C) = 1/3$, $m_\Omega(A\cup B\cup C) = 0$. This result merits some reflection. One might be surprised that even though $P(A) = .5$ is exactly known, one does not have $bel_\Omega(A) = .5$. If the frame had been A versus $\overline{A}$, the critic would have been appropriate, except that in such a frame we just have the required results. The difference observed here reflects the fact that there are three elements. What is nice is that the pignistic probability induced in this case is such that $BetP(A) = .5$ (the pignistic transformation is detailed in next section).

2) If $\mathscr{P} = \{(.5, b, c): b+c = .5\}$, then $m_\Omega(A) = 1/3$, $m_\Omega(B) = m_\Omega(C) = 0$, $m_\Omega(A\cup B) = m_\Omega(A\cup C) = 1/6$, $m_\Omega(B\cup C) = 1/4$, $m_\Omega(A\cup B\cup C) = 1/12$. The same remarks hold as for the case 1, but $BetP(A) = .5$ as it should.

3) If $\mathscr{P} = \{(a, b, c): a\leq.5, b\leq.5, c\leq.5\}$, then $m_\Omega(A) = m_\Omega(B) = m_\Omega(C) = 0$, $m_\Omega(A\cup B) = m_\Omega(A\cup C) = m_\Omega(B\cup C) = .25$, $m_\Omega(A\cup B\cup C) = .25$.

4) Suppose You know that $\mathscr{P}$ is characterized by a lower probability function $P_*$ on $\Omega$. Let $P^*$ be the upper probability function dual of $P_*$, i.e., $P^*(X) = 1 - P_*(\overline{X})$ for $X\subseteq\Omega$. Let $a = P_*(A)$, $b = P_*(B)$, $c = P_*(C)$, $A = P^*(A)$, $B = P^*(B)$, $C = P^*(C)$. The belief on $\Omega$ induced by the set $\mathscr{P}$ of probability distributions P on $\Omega$ compatible with the upper and lower probabilities (i.e., $\forall A\subseteq\Omega$, $P_*(A)\leq P(A)\leq P^*(A)$) is given by:

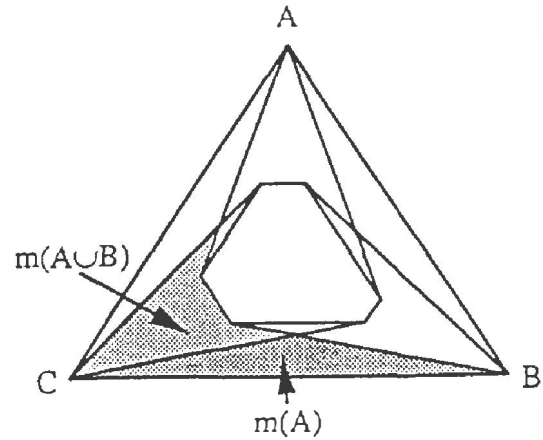

Figure 5: Domain of $P\Omega$ when $|\Omega| = 3$. The hexagon represents the set $\mathscr{P}$ of probability functions compatible with a given lower probability function. The values of $m_\Omega(A|\mathscr{P})$ and $m_\Omega(A\cup B|\mathscr{P})$ are the shaded surfaces. $m_\Omega(A\cup B\cup C|\mathscr{P})$ is the surface of the hexagon plus the three small left over triangles fixed on its side.

$$bel_\Omega(A) = \frac{a}{a+B+C}$$
$$bel_\Omega(B) = \frac{b}{A+b+C}$$
$$bel_\Omega(C) = \frac{c}{A+B+c}$$
$$bel_\Omega(A\cup B) = (1-C)^2 + C(a+b)$$
$$bel_\Omega(A\cup C) = (1-B)^2 + B(a+c)$$
$$bel_\Omega(B\cup C) = (1-A)^2 + A(b+c)$$
$$bel_\Omega(A\cup B\cup C) = 1$$

These results are obtained by computing the various surfaces described in figure 5.

It is worth noticing that bel is not equal to $P_*$, even when $P_*$ is a belief function. Why should they? The transferable belief model never requires that the belief function that quantifies our belief should be the lower envelop of a set of probability function.

## 5. PIGNISTIC PROBABILITY.

In Smets (1990a, 1993b) and Smets and Kennes (1994), we have shown how to build the appropriate probability function BetP, called the pignistic probability function, from a belief function when a decision must be made. We have shown that the only 'rational' transformation, called the pignistic transformation, must satisfy the following rule when the betting frame $\Omega$ is finite. Let m be the basic belief assignment quantifying the agent's beliefs over $\Omega$. For $\omega \in \Omega$,

$$BetP(\omega) = \sum_{A:\omega \in A \subseteq \Omega} \frac{1}{|A|} m(A)$$

where $|A|$ is the number of elements of $\Omega$ in A. Any other probability function would lead to irrationality in the betting behavior of the agent. Its extension to continuous cases is easy to realize if the bbd are really densities, in which case sums become classical integrals.

We study how decision should be made when beliefs are induced by a set of probabilities, i.e., how to derive the appropriate pignistic probability from the initial belief induced by the knowledge that $P \in \mathscr{F} \subseteq \mathbb{P}_\Omega$. The choice of the appropriate betting frame is important. We could think to build the belief function over $\Omega$ that quantifies our belief over $\Omega$ and apply the pignistic transformation to such a belief function over $\Omega$ using $\Omega$ as the betting frame. But this is an erroneous strategy as the betting frame is not $\Omega$ but W. The beliefs induced by $P \in \mathscr{F}$ is a belief over W, the belief derived on $\Omega$ is only the result of the marginalization of the first one on $\Omega$.

Using W as the betting frame, we apply the pignistic transformation to $Bel_W$. For $X \subseteq \mathbb{P}_\Omega$ let $S(X)$ be the surface of X. Suppose the agent who wants to bet on $\Omega$ knows only that $P \in \mathscr{F}$. The pignistic transformation implies that the bbd $m_W(\mathscr{F}_A(q))$ given to $\mathscr{F}(q)$ (see (4.2)) be equally distributed among the elements of $\mathscr{F}$.

$$BetP_\Omega(A) = \int_{q \in \mathbb{P}_\Omega} \frac{S(\mathscr{F}_A(q) \cap \mathscr{F})}{S(\mathscr{F})} dq.$$

Interchanging the order of integration, one gets that

$$BetP_\Omega(A) = \frac{1}{S(\mathscr{F})} \int_{P \in \mathscr{F}} \int_{q \in \mathbb{P}_\Omega} I(\mathscr{F}_A(q) \cap \{P\}) \, dq \, dP$$

where $I(X) = 1$ if $X \neq \emptyset$, 0 otherwise.

One has: $\int_{q \in \mathbb{P}_\Omega} I(\mathscr{F}_A(q) \cap \{P\}) \, dq = P(A)$,

hence:

$$BetP_\Omega(A) = \frac{1}{S(\mathscr{F})} \int_{P \in \mathscr{F}} P(A) \, dP.$$

The pignistic probability $BetP_\Omega(A)$ so derived is equivalent to the probability one would derive by assuming an equi a priori density over $\mathbb{P}_\Omega$, conditioning it on $\mathscr{F}$, and computing the expected probability of $P(A)$.

In particular, when $|\Omega| = 2$ and $\mathscr{F} = [a, b]$, the result is:

$$BetP(S) = \frac{1 + a - b}{2}.$$

These are quite natural results. $BetP_\Omega(. | \mathscr{F})$ indeed happens to be the center of gravity of $\mathscr{F}$, but its derivation does not result from the use of an equi a priori density over $\mathbb{P}_\Omega$. It just happens that both approaches lead to the same results: 1) the equi a priori density over $\mathbb{P}_\Omega$ and 2) the application of the pignistic transformation combined with the evaluation of $BetP_\Omega(X | \mathscr{F})$ as $BetP_W(cyl(X) | \mathscr{F})$ for $X \subseteq \Omega$, where $BetP_W$ is the pignistic probability obtained from $bel_W(. | \mathscr{F})$ over the betting frame W.

## 6. CONCLUSIONS:

1) Generalization to $|\Omega| > 3$ is conceptually easy, but very laborious when solutions must be written down. Nothing new comes out of it. In practice, computation will not been based on the explicit equations, but on some Monte Carlo method.

2) Generalization of the procedure can be achieved if one has a non-degenerated belief function on $\mathbb{P}_\Omega$ if there are only a finite number of subsets of $\mathbb{P}_\Omega$ that receive positive basic belief masses (more general cases are not considered here). Let $\{\mathscr{F}_i: i = 1, 2...n\}$ be the set of focal elements of $bel_{\mathbb{P}_\Omega}$ with their basic belief masses $m_{\mathbb{P}_\Omega}(\mathscr{F}_i)$. For each focal element $\mathscr{F}_i$, we derive $bel_W(. | P \in \mathscr{F}_i)$ over W. The belief function $bel_W$ over $\Omega$ induced $\{(\mathscr{F}_i, m_{\mathbb{P}_\Omega}(\mathscr{F}_i)): i = 1, 2...n\}$ is:

$$\forall A \subseteq W \quad bel_W(A) = \sum_{i=1}^{n} bel_W(A | P \in \mathscr{F}_i) \, m_{\mathbb{P}_\Omega}(\mathscr{F}_i)$$

3) Suppose two pieces of evidence that say that $P \in \mathscr{F}_1$ and $P \in \mathscr{F}_2$, respectively. The combination of these two pieces of evidence leads to the knowledge $P \in \mathscr{F}_1 \cap \mathscr{F}_2$.

One could build $bel_1$ on W as the belief function induced by the knowledge that $P \in \mathscr{F}_1$. Identically, one could build $bel_2$ on W as the belief function induced by the knowledge that $P \in \mathscr{F}_2$. One could then be tempted, erroneously in fact, to combine $bel_1$ and $bel_2$ into $bel_1 \oplus bel_2$ by Dempster's rule of combination.

One could also build $bel_{12}$ on W as the belief function induced by the knowledge that $P \in \mathscr{F}_1 \cap \mathscr{F}_2$. In general $bel_{12} \neq bel_1 \oplus bel_2$. Only $bel_{12}$ is correct. Indeed Dempster's rule of combination is applicable iff both pieces of evidence are distinct, and distinctness is not satisfied in the present context because of the existence of a unique underlying probability function on $\Omega$ that create a link between the two pieces of evidence.

4) In conclusion, the knowledge that the probability function P over $\Omega$ belongs to some subset $\mathscr{F}$ of $\mathbb{P}_\Omega$ permits the construction of a belief function bel over $\mathbb{P}_\Omega \times \Omega$ and over $\Omega$. It must be enhanced that in general the belief function $bel_\Omega$ induced over $\Omega$ by a lower probability function $P_*$ will not satisfy $bel_\Omega = P_*$ even if $P_*$ happens to be a belief function. By showing what is the belief induced by a lower probability, we hope we have been able to show the fundamental difference between the upper and lower probabilities model and the transferable belief model (see also Smets, 1987, Smets and Kennes, 1994, Halpern and Fagin, 1990).

Acknowledgment: Research work has been partly supported by the Action de Recherches Concertées BELON funded by a grant from the Communauté Française de Belgique and the ESPRIT III, Basic research Action 6156 (DRUMS II) funded by a grant from the Commission of the European Communities.

# Ignorance and the Expressiveness of Single- and Set-Valued Probability Models of Belief

**Paul Snow**
P.O. Box 6134
Concord, NH 03303-6134 USA
paulsnow@delphi.com

## Abstract

Over time, there have been refinements in the way that probability distributions are used for representing beliefs. Models which rely on single probability distributions depict a complete ordering among the propositions of interest, yet human beliefs are sometimes not completely ordered. Non-singleton sets of probability distributions can represent partially ordered beliefs. Convex sets are particularly convenient and expressive, but it is known that there are reasonable patterns of belief whose faithful representation require less restrictive sets. The present paper shows that prior ignorance about three or more exclusive alternatives and the emergence of partially ordered beliefs when evidence is obtained defy representation by any single set of distributions, but yield to a representation based on several sets. The partial order is shown to be a partial qualitative probability which shares some intuitively appealing attributes with probability distributions.

## 1. INTRODUCTION

Probability distributions have long been advocated as a useful foundation for the modeling of beliefs. The best known form of probabilistic belief representation consists of a single distribution. Such models bring with them a well-developed normative theory of behavior in the face of risk (Savage, 1972) which has had many adherents over the years.

Recently, some researchers have concluded that single distribution models are too restrictive. Beliefs may not always be completely ordered by the believer, even though a single probability distribution necessarily represents them as being so. Nevertheless, other attributes of probability distributions do seem like accurate portrayals of how beliefs behave with respect to Boolean combinations of the underlying events, and of how beliefs change in the face of evidence. Some of these desirable attributes are peculiar to probability distributions. So, to have the attributes, a belief representation must either use probability distributions or else use measures that agree with some probability distributions (Snow, 1992).

One way to get the desirable attributes of probabilities without the undesirable restrictiveness of a complete ordering is to model beliefs using non-singleton sets of probability distributions. It is often convenient to use convex sets of probability distributions, which arise as solutions to systems of simultaneous linear inequalities. Many natural language expressions of belief are easily translated into linear inequality constraints (Nilsson, 1986), e.g. "This event is at least as likely as that one." Linear constraint systems can be revised simply by Bayes' formula (Snow, 1991). Although there is a diversity of opinion about how set estimates might inform decision making, there are useful suggestions for decision rules in the literature (for a review, see Sterling and Morrell, 1991).

As versatile as convex sets are, there are reasonable belief patterns that convex sets fail to represent. For example, the set of posterior probabilities derived from a convex set of priors and a convex set of conditionals is generally not convex (White, 1986). Further, some important constraints are non-linear. Kyburg and Pittarelli (1992) discuss the non-convex sets which arise from the non-linear assumption of independence between events.

The present paper explores a circumstance where no single set of probability distributions, convex or otherwise, faithfully represents a reasonable pattern of belief, namely, ignorance being overcome by evidence when there are more than two alternatives. By *ignorance*, we mean that the believer is unwilling to assert any non-trivial prior ordering among the sentences of interest. By *being overcome by evidence*, we mean that the believer will assert some non-trivial orderings if the contrast between the conditional probabilities for the evidence given the sentences is sufficiently impressive.

A probabilistic solution to the representation of ignorance being overcome by evidence is presented. Although the model is more complex than a single set of probability distributions, the orderings that arise have much in common with single posterior probability distributions, and inference about the orderings is computationally inexpensive.

## 2. NOTATION AND ASSUMPTIONS ABOUT IGNORANCE

In this paper, we shall use the notation

$$S >e> T$$

to denote the condition that the believer asserts that sentence $S$ is, with a warrant satisfactory to the believer, at least as belief-worthy as sentence $T$ in light of evidence $e$. If evidence $e$ does not lead the believer to assert an ordering of sentences $S$ and $T$, then we write

$$S ?e? T$$

The condition of having no relevant evidence is indicated by the particle *nil*, as in

$$S ?nil? T$$

which expression denotes that there is no ordering between some sentences $S$ and $T$ in the absence of evidence.

We shall assume that the sentences of interest belong to a partitioned domain, which is defined as follows:

**Definition.** A *partitioned domain* is a set comprising:

(i) the always-true sentence, denoted **true**

(ii) the always false sentence, denoted **false**

(iii) two or more mutually exclusive sentences, called *atoms*

(iv) well-formed expressions involving atoms, **or**, and parentheses, called *simple disjunctions*

(v) well-formed expressions involving simple disjunctions, **true, false, or, not,** and parentheses

We shall assume throughout that the atoms in the domain are collectively exhaustive, that is, one of the atoms is true. This additional assumption places little epistemological burden on the believer (at worst, it means that one of the atoms is "none of the other atoms are true"), and has the convenient effect that every sentence in the domain has an equivalent simple disjunction. Finally, although infinite domains are useful in such applications as statistical hypothesis testing, we shall assume throughout this paper that the number of atoms in the domain is finite.

Our first assumptions about ignorance, and the conquest of ignorance by evidence express the following ideas. If no evidence has yet been observed, and the question of relative belief-worthiness is not answerable on logical grounds, then there is no satisfactory warrant to order one sentence ahead of another. Even after evidence has been observed, the question may remain open. Once a commitment to an ordering is made, then other commitments may be inferred by conditional probability considerations. A belief-ordering consistency principle discussed by Sugeno (unpublished dissertation, cited in Prade, 1985) obtains regardless of the presence or absence of evidence. The formal assumptions are:

**A1.** (Lack of explicit non-trivial prior orderings) For any sentences S and T,

$$S >nil> T \text{ implies that T implies S.}$$

**A2.** (Lack of implicit non-trivial prior orderings) Values for conditional probabilities and orderings among them are neither known nor assumed if those values or orderings imply non-trivial constraints on the prior probabilities.

**A3.** (Consistency) For all evidence e, including *nil*, and any sentences S and T,

$$\text{if T implies S, then } S >e> T.$$

**A4.** (Impartiality) If $S >e> T$, and S' and T' are sentences, and S is exclusive of T, then

if S' is exclusive of T and $p(e|S') \geq p(e|S)$, then $S' >e> T$, and

if S is exclusive of T' and $p(e|T) \geq p(e|T')$, then $S >e> T'$.

**A5.** (Recovery from ignorance about atoms) For exclusive atoms s and t, and non-nil evidence e, a necessary condition for $s >e> t$ is that $p(e|s) \geq p(e|t)$, and if $p(e|s) > 0$, then the inequality is strict. If $p(e|s) > 0$, then $p(e|t) = 0$ is not a necessary condition for $s >e> t$.

**A6.** (Dominance) For any sentences S, T, U and U' where (S and U) and (T and U) are both false and U' implies U, and for all evidence e, including *nil*,

if $(S \text{ or } U') >e> (T \text{ or } U)$, then $S >e> T$, and

if $(S >e> T)$, then $(S \text{ or } U) >e> (T \text{ or } U')$.

## 3. COMMENTARY ON THE ASSUMPTIONS

Assumption A1 explains one circumstance where we decline to assert any ordering: when there is no evidence, and the one sentence doesn't imply the other. A2 restricts the scope of the assumptions to problems whose givens rule out no prior probability distribution over the atoms. The conditions in assumption A2 reflect the easily-shown fact that a disjunctive conditional like $p(e|S)$ is a convex combination of the conditionals for the atoms in $S$, with weights proportional to the prior probabilities of the atoms.

Assumption A3 says that we always assert an ordering